\def\year{2022}\relax
\documentclass[letterpaper]{article} 
\usepackage{aaai22}  
\usepackage{times}  
\usepackage{helvet}  
\usepackage{courier}  
\usepackage[hyphens]{url}  
\usepackage{graphicx} 
\usepackage{natbib}  
\usepackage{caption} 
\DeclareCaptionStyle{ruled}{labelfont=normalfont,labelsep=colon,strut=off} 
\usepackage{epsfig}
\usepackage{amsmath}
\usepackage{amssymb}

\usepackage{multirow}
\usepackage{booktabs}
\usepackage{enumerate}
\usepackage{subfig}

\usepackage{algorithm}
\usepackage{algorithmic}

\usepackage{newfloat}
\usepackage{listings}
\floatstyle{ruled}
\newfloat{listing}{tb}{lst}{}
\floatname{listing}{Listing}
\nocopyright
\title{DRAG: Dynamic Region-Aware GCN for Privacy-Leaking Image Detection}
\author {
    Guang Yang\textsuperscript{\rm 1,2},
    Juan Cao\textsuperscript{\rm 1,2},
    Qiang Sheng\textsuperscript{\rm 1,2},
    Peng Qi\textsuperscript{\rm 1,2},
    Xirong Li\textsuperscript{\rm 3},
    Jintao Li\textsuperscript{\rm 1}
}
\affiliations {
    \textsuperscript{\rm 1} Key Laboratory of Intelligent Information Processing, \\Institute of Computing Technology, Chinese Academy of Sciences\\
    \textsuperscript{\rm 2} University of Chinese Academy of Sciences\\
    \textsuperscript{\rm 3} Key Lab of Data Engineering and Knowledge Engineering, Renmin University of China\\
    $\{$gyang, caojuan, shengqiang18z, qipeng, jtli$\}$@ict.ac.cn, xirong@ruc.edu.cn
}

\usepackage{bibentry}

\begin{document}

\maketitle

\begin{abstract}
The daily practice of sharing images on social media raises a severe issue about privacy leakage. To address the issue, privacy-leaking image detection is studied recently, with the goal to automatically identify images that may leak privacy.
Recent advance on this task benefits from focusing on crucial objects via pretrained object detectors and modeling their correlation. However, these methods have two limitations: 1) they neglect other important elements like scenes, textures, and objects beyond the capacity of pretrained object detectors; 2) the correlation among objects is fixed, but a fixed correlation is not appropriate for all the images.
To overcome the limitations, we propose the \textbf{D}ynamic \textbf{R}egion-\textbf{A}ware \textbf{G}raph Convolutional Network (\textbf{DRAG}) that dynamically finds out crucial regions including objects and other important elements, and models their correlation adaptively for each input image. To find out crucial regions, we cluster spatially-correlated feature channels into several region-aware feature maps. Further, we dynamically model the correlation with the self-attention mechanism and explore the interaction among the regions with a graph convolutional network. The DRAG achieved an accuracy of 87\% on the largest dataset for privacy-leaking image detection, which is 10 percentage points higher than the state of the art. The further case study demonstrates that it found out crucial regions containing not only objects but other important elements like textures. The code and more details are in \url{https://github.com/guang-yanng/DRAG}.

\end{abstract}

\section{Introduction}
\label{intro}
 
\begin{figure}[!t]
\centering
\includegraphics[width=1\columnwidth]{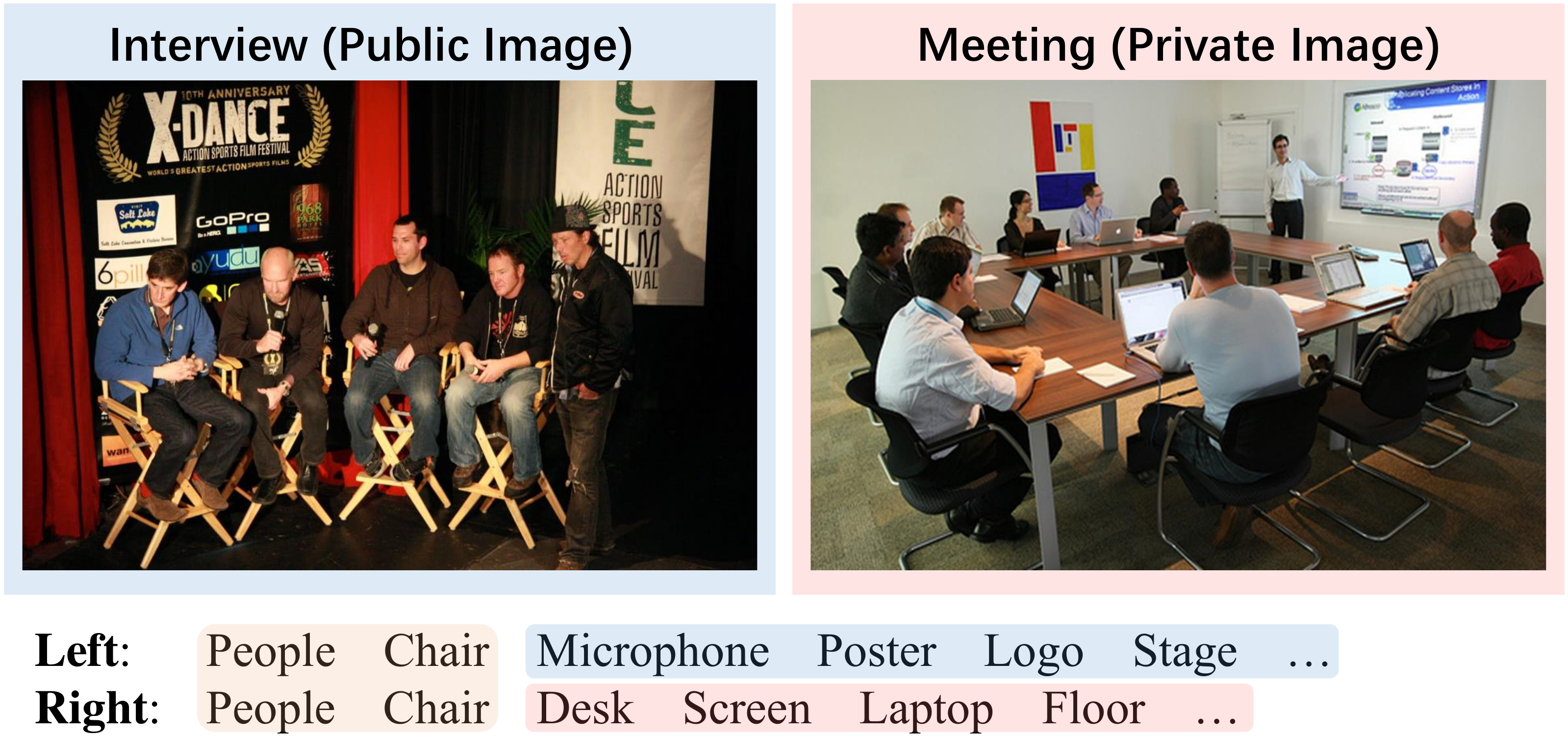}
\caption{Example of images that people share online in the Image Privacy dataset~\cite{yang2020graph}. (a) is a public image that is safe for sharing, while (b) is a private image that may leak sensitive information. Public and private images contain many common elements as well as specific ones. The co-occurrence and interaction of the elements provide semantic clues of scenes, activities, etc., which is crucial for privacy-leaking image detection. Therefore, methods for this task need to find out the elements and take their co-occurrence and interaction into consideration.
}
\label{privacy_cases}
\end{figure}

Social media like Facebook has been part of our daily life. People post a large number of images on social media to record and share their lives.
However, the convenience of online image sharing brings about the risk of privacy leakage. 
The shared images contain rich information like personal relationships and physical disability~\cite{orekondy2017towards}. 
Malicious use of such information has been documented~\cite{deepfake} causing dire consequences like fraud and cyber violence~\cite{fraud}.
As a severe issue that is close to our daily life, privacy-leaking image is attracting increasing concerns. 

Social media platforms allow users to set privacy preferences like the visibility of their content to protect privacy, but many users still unconsciously share images that may leak privacy.
Although people have common expectations about the privacy setting of online images~\cite{hoyle2020privacy}, they often
lack the awareness of the privacy risk of the shared images~\cite{tuunainen2009users, wang2011regretted}. \citet{liu2011analyzing} proved that there is a gap between users' expectations and the reality of users' privacy settings of the shared images.
The above phenomenon and the potential harms make it urgent to help users reduce the privacy risks during image sharing.
Users may unintentionally share images that leak privacy and the spread of the images is almost uncontrollable. 
Therefore, a feasible method to reduce the privacy risks is to automatically identify images that may leak privacy and give warnings to users before sharing.

We use \textit{private images} to refer to images that may leak privacy and \textit{public images} to refer to images that are safe for sharing. Researchers mainly consider non-personalized consensus and build corresponding datasets, and several examples are presented in Fig.~\ref{privacy_cases}. 
Following~\citet{tran2016privacy} and~\citet{yang2020graph}, 
we formulate privacy-leaking image detection as a binary classification task (i.e., predict whether a given image is \textit{private} or \textit{public}).
Fig.~\ref{privacy_cases} demonstrates that the interaction among the elements in the images provides clues and helps distinguish between private and public images.~\citeauthor{yang2020graph} focuses on objects and their correlation to identify private images based on object detection. 
However, they neglect other important elements like scenes~\cite{tonge2019dynamic}, textures, and objects beyond the capacity of pre-trained object detectors.
Furthermore, the correlation among objects is fixed, but the elements vary in different images, making the fixed correlation inappropriate.

\begin{figure*}[!t]
\centering
\includegraphics[width=2\columnwidth]{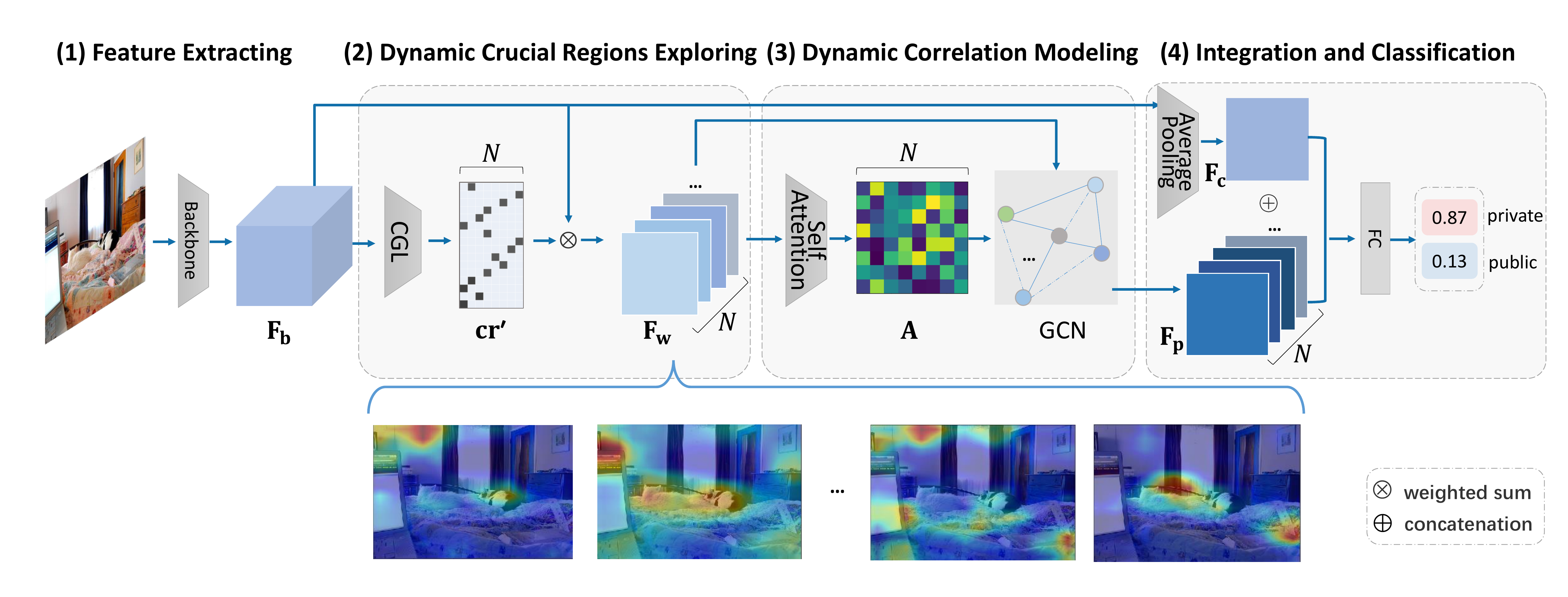}
\caption{Workflow of the Dynamic Region-Aware GCN (DRAG) for privacy-leaking image detection. 
(1) DRAG first extracts the feature $\mathbf{F_b}$ of the input image. (2) The channels of $\mathbf{F_b}$ are then clustered into $N$ groups with Channel Grouping Layer ($CGL$). According to the approximate clustering result $\mathbf{cr^\prime}$, $\mathbf{F_b}$ are aggregated into $N$ feature maps $\mathbf{F_w}$ to represent $N$ differentiated regions (Examples are at the bottom). (3) DRAG formulates a graph with $\mathbf{F_w}$ as the $N$ nodes and uses the self-attention mechanism to obtain the correlation matrix $\mathbf{A}$. Then a GCN is used for feature learning on this graph.
(4) The learned feature $\mathbf{F_p}$ is concatenated with a global representation of the image $\mathbf{F_c}$ to identify private images.}
\label{workflow}
\end{figure*}

To overcome the limitations, we propose  \textbf{D}ynamic \textbf{R}egion-\textbf{A}ware \textbf{G}raph Convolutional Network (\textbf{DRAG}) to dynamically find out regions of the crucial elements, and model their correlation adaptively per image. The workflow of DRAG is presented in Fig.~\ref{workflow}, which contains two main parts. In the first part (Fig.~\ref{workflow} (2)), the DRAG finds out $N$ crucial regions from the feature map without the reliance on the object detectors. Specifically, based on the feature map obtained from the backbone, the DRAG clusters the spatially-correlated feature channels into $N$ region-aware feature maps. In the second part (Fig.~\ref{workflow} (3)), DRAG adopts the graph convolutional network (GCN) to model the interaction among the $N$ regions. 
The regions are obtained dynamically for each image, and thus the correlation among the regions should be adaptive rather than predefined and fixed. We dynamically model the correlation with the self-attention mechanism to initialize the correlation matrix for GCN.
Then the interaction among the $N$ crucial regions is explored by propagating corresponding features through GCN with the adaptive correlation matrix.
Finally (Fig.~\ref{workflow} (4)), the propagated features are concatenated with a global representation of the image to identify private images. 
Compared with existing works, the dynamic nature of the DRAG enables it to find out more diverse elements (not only objects) and model their correlations adaptively. 

Our main contributions are summarized as follows:
\begin{enumerate}[(1)]
    \item We proposed a novel framework DRAG for privacy-leaking image detection. The DRAG dynamically finds out crucial regions without the limitation of pretrained object detectors and models the correlation among crucial regions adaptively for each image.
    \item To explore the interaction among the crucial regions, we proposed a region-aware method to initialize the graph for GCN based on spatially-correlated channels clustering and self-attention mechanism.
    \item The experimental results prove the effectiveness of the proposed framework for privacy-leaking image detection. 
    The DRAG that only utilizes visual features outperformed existing methods, including visual-based and multi-modal ones.
\end{enumerate}


\section{Related Work}

\subsection{Privacy-Leaking Image in Online Image Sharing}
\citet{liu2020privacy} concluded several privacy issues of online image sharing. In this paper, we focus on the unawareness of privacy during image sharing. 
There are two main types of methods to deal with the risk of unawareness of privacy.

The first type of method mainly adopts classification models to identify private images.
\citet{zerr2012picalert} proposed a privacy-aware classifier based on visual features like face and color histograms.~\citet{buschek2015automatic} proposed a multi-modal method that assigns privacy labels to the images based on visual features and metadata like location and publication time. 
\citet{tonge2018uncovering} utilized another kind of metadata, tag, and~\citet{tonge2019dynamic} further derived features of the object, scene, and tags for privacy-leaking image detection.~\citet{yang2020graph} extracted a knowledge graph from the images and identified private images based on object detection and graph neural networks.

The second type of method focuses on sensitive regions in the images, including approaches like object detection and semantic segmentation.
Some detected private attributes such as faces~\cite{sun2018face}, license plates~\cite{zhou2012principal}, and social relationship~\cite{li2017dual}. 
\citet{orekondy2017towards} defined a list of privacy attributes and detected them simultaneously. 
Some works attempted to protect privacy-leaking image based on blurring~\cite{fan2018image}, blocking~\cite{li2017blur}, cartooning~\cite{hasan2017cartooning}, and perturbation~\cite{oh2017adversarial}.~\citet{shetty2018adversarial} removed private objects from the images based on a generative method. However, a person may be recognized even his face is not visible~\cite{oh2016faceless}, and the redacted image may be recovered 
~\cite{shen2019single}. 
As the usage of shared images is almost uncontrollable, it is better to prevent the risk from the beginning. Therefore, we follow the first type of method to solve the issue of privacy-leaking images by classification.

\subsection{Graph-based Methods in Visual Tasks}

Graph-based methods have shown great potential in many vision tasks in recent years, including visual question answering~\cite{teney2017graph}, person re-identification~\cite{wu2019unsupervised}, multi-label image recognition~\cite{marino2016more}, and relationship recognition~\cite{wang2018deep}.
\citet{yeattention} utilized GCN~\cite{kipf2016semi} for multi-label classification.~\citet{yang2020graph} adopted graph neural networks for privacy-leaking image detection and showed that modeling the interaction among crucial elements is an effective way. However, their framework only focuses on the objects that the pretrained object detector can recognize.
Inspired by \citet{yeattention} and ~\citet{yang2020graph}, we proposed DRAG that can model the interaction among more crucial elements dynamically with GCN. Furthermore, instead of a fixed correlation matrix for all the images~\cite{yang2020graph}, we extract the correlation matrix for each input image adaptively based on the self-attention mechanism.

\section{Approach}

\subsection{Overview of DRAG}
The DRAG dynamically finds out crucial regions and models their correlation adaptively for each input image.
Then the DRAG explores the interaction among the crucial regions with GCN and identifies the private images. 

Specifically, the DRAG contains two main parts (see Fig.~\ref{workflow}). In the first part (Fig.~\ref{workflow}  (2)), the DRAG extracts diverse and tiny clues and then clusters them to obtain region-aware feature maps as the representation of crucial regions, without the reliance on the object detectors. In the second part (Fig.~\ref{workflow} (3)), the DRAG models the correlation among the crucial regions based on the self-attention mechanism and initializes the correlation matrix for GCN. Then the features of crucial regions are propagated through GCN with the adaptive correlation matrix to explore the interaction among these regions. Finally (Fig.~\ref{workflow} (4)), the propagated features are concatenated with a global representation of the image to classify a given image as private or public. 

\subsection{Dynamic Crucial Regions Exploring}
\label{CGL}
Tasks like object detection and fine-grained image recognition need to focus on objects for better performance. For example, the Region Proposal Network~\cite{ren2015faster} is used to select regions that may contain objects, attention-based methods~\cite{fu2017look} are used to focus on the details of objects for fine-grained image classification. However, the clues for privacy-leaking image detection are revealed not only by the objects but also by other elements such as scenes and textures
. To focus on these crucial elements, we find out differentiated regions in an image based on the channel grouping mechanism~\cite{zheng2017learning}. 

We first trained a backbone (here, ResNet~\cite{he2016deep}) and got the convolutional feature of the input image $\mathbf{F_b} \in \mathcal{R}^{C\times H\times W}$, where $W$, $H$, and $C$ are the width, height, and channel number of the feature. According to~\citet{simon2015neural}, the peak responses of the channels correspond to various visual patterns. Following~\citet{zheng2017learning}, we clustered the channels by K-means~\cite{macqueen1967some} according to spatial correlation among the corresponding peak responses and adopted the cluster results as the representation of crucial regions.
To combine the clustering with neural networks, the clustering process was approximated by several fully-connected layers ($FCs$), and the details are as follows:

A group of channels whose peak responses appear in neighboring locations were clustered together. For each channel, we got the coordinates of the peak response $\left[t_{x}, t_{y}\right]$ of all training images and formulated them into a vector:
$\left[t_{x}^{1}, t_{y}^{1}, t_{x}^{2}, t_{y}^{2}, \ldots, t_{x}^{\Omega}, t_{y}^{\Omega}\right]$, where $t_{x}^{i}$ and $ t_{y}^{i}$ are the coordinates of peak response of the $i^{th}$ training image, and $\Omega$ refers to the size of the training set. This vector was used as the feature for clustering with K-means. The channels were clustered into $N$ groups to represent $N$ differentiated regions. The clustering results were formulated as a matrix $\mathbf{cr} \in \mathcal{R}^{N\times C}$ with $cr_{ij}=1$ or $0$, which indicates that if the $j^{th}$ channel belongs to the $i^{th}$ group (i.e., the $i^{th}$ region).

The pretrained backbone initially focuses on the objects and will be fine-tuned to adapt to privacy-leaking image detection.
To let the clustering results obtained from the backbone be optimized together, we adopted $FCs$ to approximate the clustering process, which is called the channel grouping layer ($CGL$). $CGL$ takes the feature map $\mathbf{F_b}$ as input, then outputs the estimated result of clustering $\mathbf{cr ^{\prime}}\in \mathcal{R}^{N\times C}$:
\begin{equation}
\mathbf{cr ^{\prime}}=CGL(\mathbf{F_b})=sigmoid(FCs(\mathbf{F_b})),
\end{equation}

We used the $\mathbf{cr ^{\prime}}$ to get the feature of each region with an averaged weighted sum mechanism. For the $i^{th}$ region, its corresponding feature $\mathbf{F_{wi}} \in \mathcal{R}^{H\times W}$ was obtained by :
\begin{equation}
\mathbf{F_{wi}}=\frac{1}{C}\sum_{c}\mathbf{F_{bc}}*cr ^{\prime}_{ic},
\end{equation}
where $ C$ is the number of channels in $\mathbf{F_b}$, $\mathbf{F_{bc}}$ is the feature of the $c^{th}$ channel. $\mathbf{cr ^{\prime}_{ic}}$ is the estimated indicator that if the $c^{th}$ channel belongs to the $i^{th}$ region. By concatenating features of all regions, we finally obtained the $\mathbf{F_w} \in \mathcal{R}^{N \times H\times W}$.

\noindent \textbf{Initialization} To obtain a proper initialization, $CGL$ was pretrained to let the $\mathbf{cr ^{\prime}}$ be as close to $\mathbf{cr}$ as possible, and the details are described in the Experiments section.
During the joint learning, we adopted two losses, $Dis(\cdot)$ and $Div(\cdot)$, to force the $CGL$ to learn differentiated regions as follows:
\begin{equation}
\small
\begin{aligned}
Dis \!\left(\mathbf{F_w}\right)\!&=\!\sum_{i\in N}\!\sum_{(x, y)\!\in \!\mathbf{r_i}}\! \mathbf{F_{wi}}(x\!,\! y)^2\!\left[\left\|x\!-\!t_{ix}\right\|^{2}\!+\!\left\|y\!-\!t_{iy}\right\|^{2}\right],\\
Div \!\left(\mathbf{F_{w}}\right)\!&=\!\sum_{i\in N}\!\sum_{(x, y)\!\in \!\mathbf{r_i}}\!\mathbf{F_{wi}}(x\!,\! y)^2\left[\max _{j \neq i} \mathbf{F_{wj}}(x\!,\! y)\!-\!mrg\right]^2\!,\end{aligned}
\end{equation}
where $i$ refers to the $i^{th}$ region, $t_{ix}$ and $t_{iy}$ are the coordinates of peak response in the $i^{th}$ region. $mrg$ is the mean of all the values in feature map $\mathbf{F_w}$, which represents a margin to make $Div(\cdot)$ less sensitive to noise. The $Dis(\cdot)$ encourages a compact distribution in the feature of a region (i.e., similar visual patterns from a specific part to be grouped together), while the $Div(\cdot)$ forces the model to learn diverse regions rather than similar ones. Such constraints make the $CGL$ learn differentiated regions for image privacy detection.

\subsection{Dynamic Correlation Modeling}

We obtained the feature of several crucial regions $\mathbf{F_w}$ based on $CGL$. To explore the interaction among these crucial regions for privacy-leaking image detection, we
formulated a graph with regions as nodes and correlation among the regions as the edges to take the advantage of GCN. 
The regions were obtained dynamically for each image, and thus the correlation should be adaptive rather than predefined and fixed.
Inspired by~\citet{yeattention} and~\citet{vaswani2017attention}, we proposed a dynamic way to get an adaptive correlation matrix for GCN.

To model the correlation among the $N$ crucial regions, we adopted the self-attention mechanism~\cite{vaswani2017attention} which is widely used in NLP tasks to learn the correlation among words.
We first got three vectors query ($q$), key ($k$), and value ($v$) from $\mathbf{F_{wi}}$ for each region $r_i$ with three fully-connected layers. For all the $N$ regions, the matrices $ \mathbf{Q\in \mathcal{R}^{N\times d_k}}$, $ \mathbf{K}\in \mathcal{R}^{N\times d_k}$ and $ \mathbf{V}\in \mathcal{R}^{N\times N}$ were calculated by:
\begin{equation}
\begin{aligned}
\mathbf{Q}\!&=\!\mathbf{W_q} \mathbf{F}_w\!+\!\mathbf{b_q},
\mathbf{K}\!&=\!\mathbf{W_k} \mathbf{F}_w\!+\!\mathbf{b_k},
\mathbf{V}\!&=\!\mathbf{W_v} \mathbf{F}_w\!+\!\mathbf{b_v},
\end{aligned}
\end{equation}
where $d_k$ is the dimension of both $q$ and $k$, and $\mathbf{W}$ and $\mathbf{b}$ refer to the weights and biases in fully-connected layers, respectively. The results of self-attention $\mathbf{A}$ was given by:
\begin{equation}
\mathbf{A}=Attention(\mathbf{Q, K, V})=softmax\left(\frac{\mathbf{Q} \mathbf{K}^{T}}{\sqrt{d_{k}}}\right) \mathbf{V}.
\end{equation}
Each value in the matrix $\mathbf{A}\in \mathcal{R}^{N\times N}$ is obtained by considering one region and all other ones.
As a result, $\mathbf{A}$ is able to represent the correlation among the $N$ regions.

\subsection{Feature Integration and Classification}

We adopted GCN with the activation function of $ReLU$ to explore the interaction among the crucial regions, which propagates features through the nodes as follows:
\begin{equation}
GCN(\mathbf{X})=ReLU(\hat{\mathbf{D}}^{-1 / 2} \hat{\mathbf{A}} \hat{\mathbf{D}}^{-1 / 2} \mathbf{X} \Theta),
\end{equation}
where $\mathbf{I}$ refers to an identical matrix, $\hat{\mathbf{A}}=\mathbf{A}+\mathbf{I}$ denotes the adjacency matrix with inserted self-loops, $\hat{D}_{i i}=\sum_{j=0} \hat{A}_{i j}$ is the diagonal degree matrix, and $\Theta$ is the learned weights. To avoid over-smoothing of node features, we only adopted two GCN layers and finally got the propagated feature $\mathbf{F_p} \in \mathcal{R}^{N\times H \times W}$:
\begin{equation}
\mathbf{F_p}=GCN(GCN(\mathbf{F_w})).
\end{equation}

To prevent that the learned regions neglect important global information in the image, we got a compressed feature $\mathbf{F_c}\in \mathcal{R}^{1\times H\times W}$ from the original feature map  $\mathbf{F_b}$ by average $\mathbf{F_b}$ through the channels. At last, the $\mathbf{F_c}$ was concatenated with $\mathbf{F_p}$ for classification with a fully-connected layer $FC$ and the activation function of $softmax$:
\begin{equation}
{\hat{y}} =softmax(FC(\mathbf{F_c} \oplus \mathbf{F_p})),
\label{softmax}
\end{equation}
where $\oplus$ denotes the concatenating operation. The output $\hat{y}$ represents the probability that if the input image is private.

\section{Experiments}

\subsection{Experimental Setup}
\subsubsection{Datasets}
\label{datasets}
\ \\
\indent \textbf{PicAlert} PicAlert~\cite{zerr2012privacy} is the first dataset for privacy-leaking image detection on social media, which was built on an average community notion of privacy. They first crawled images from image-sharing social media Flickr\footnote{https://www.flickr.com/}, then asked external viewers to judge the privacy of the photos via a social annotation game. After removing invalid annotations, they finally proposed a dataset of images with user-classified privacy labels. The PicAlert we used contains 7,518 private images and 24,615 public images.

\textbf{Image Privacy} The PicAlert is somewhat biased as most of the private images in PicAlert is person-containing. To diversify private images, \citet{yang2020graph} extended PicAlert to include more types of images reported in the previous study~\cite{tran2016privacy}, such as driver licenses, ID cards, and legal documents. The Image Privacy dataset contains 13,910 private images and 24,615 public images.

\subsubsection{Methods for Comparison}
We compared DRAG with state-of-the-art methods, including Privacy-CNH~\cite{tran2016privacy} and GIP~\cite{yang2020graph} that only utilize visual information obtained from the images, as well as Combination of Object, Scene, and User Tags~\cite{tonge2018uncovering} and DMFP~\cite{tonge2019dynamic} that utilize extra user tags\footnote{Tags that users annotate when sharing images, often contain information that cannot be obtained from the image.}.

\textbf{Privacy-CNH }(Hereafter, PCNH)~\cite{tran2016privacy} proposed a framework that utilized both object and convolutional features for privacy-leaking image detection. 
The features are finally concatenated for classification.

\textbf{GIP}~\cite{yang2020graph} is the first to adopt graph neural networks for privacy-leaking image detection. The GIP first detects objects in an image based on Faster-RCNN~\cite{ren2015faster}, and propagates the object features through a predefined graph which is extracted from the training set.

\textbf{Combination of Object, Scene, and User Tags} (Hereafter, Combination)~\cite{tonge2018uncovering} combines object tags, scene tags, and user tags for privacy-leaking image detection, which is a basic multi-modal method. The object tags and scene tags are extracted from the visual features, while the user tags are extra collected.

\textbf{DMFP}~\cite{tonge2019dynamic} is also a multi-modal method that utilizes object features, scene features, and tag features instead of the tags. The DMFP estimates the competence of the modalities and fuses the decisions dynamically. DMFP-O and DMFP-S denote DMFP that only utilize object features and scene features, respectively.

\subsubsection{Implementation}
\label{Implementation}
We conducted experiments on the two datasets to compare with state-of-the-art methods for privacy-leaking image detection. To make a fair comparison, we adopted the same experiment settings as~\citet{tonge2019dynamic} and~\citet{yang2020graph}. The ratio of train, val, and test set is 15:7:10 in both datasets. The public and private images are in the ratio of about 3:1 in PicAlert and about 7:4 in Image Privacy. 

The models were implemented with python 3.6.8, PyTorch 1.4.0~\cite{pytorch}, torchvision 0.5.0~\cite{torchvision}, and torch-geometric 1.6.1~\cite{torch_geometric}. We first pretrained a ResNet as the backbone model. We extracted the feature outputted by the last convolutional layer and obtained $\mathbf{F_b} \in \mathcal{R}^{2048\times14\times14}$. 
We clustered the channels into different regions following the process described in the Approach section
and got the clustering result $\mathbf{cr} \in \mathcal{R}^{N\times2048}$. We experimented with several region numbers $N$, including $4, 6, 8, 10$, and $12$, to explore the influence of different region numbers. The clustering result $\mathbf{cr}$ was used to pretrain the $CGL$ to let the $\mathbf{cr ^{\prime}}$ be as close to $\mathbf{cr}$ as possible, to enable the $CGL$ to learn differentiated regions. We calculated the cross-entropy loss of all the $N$ regions and $2048$ channels to optimize the $CGL$:
\begin{equation}
\small
L_{CGL}\!=\!-\!\sum_{i\in N}\!\sum_{j\in C} \left[y_{ij} \log \left(\hat{y_{ij}}\right)\!+\!\left(1-y_{ij}\right) \log \left(1-\hat{y_{ij}}\right)\right],
\end{equation}
where $y_{ij}$ is the true label of the $j^{th}$ channel in the $i^{th}$ group, and $\hat{y_{ij}}$ is the corresponding predicted probability.

The correlation matrix $\mathbf{A}$ used for $GCN$ was learned during training with the self-attention mechanism.
For the self-attention module, the dimension of $q$ and $k$ was $64$, while the dimension of $v$ was $N$. 
For the $GCN$, the number of nodes was the same as the number of regions $N$. Feature of each region was used to initialize the corresponding node, and thus the feature of each node $i$ was $\mathbf{F_{wi}}\in \mathcal{R}^{14\times 14}$. After exploring the interaction among nodes, the $GCN$ outputted $\mathbf{F_p} \in \mathcal{R}^{N\times 14\times 14}$. Finally, by concatenating the global feature $\mathbf{ F_c}\in \mathcal{R}^{1\times 14\times 14}$,  $\mathbf{(F_c} \oplus \mathbf{F_p}) \in \mathcal{R}^{(N+1)\times 14\times 14}$ was used for classification.
For the binary classification task, we adopted cross-entropy loss function:
\begin{equation}
L_{cls}=-\sum_{i}\left[y_{i} \log \left(\hat{y_{i}}\right)+\left(1-y_{i}\right) \log \left(1-\hat{y_{i}}\right)\right],
\end{equation}
where $y_{i}$ is the true label of the $i^{th}$ sample, and $\hat{y_{i}}$ is the corresponding predicted probability given by the model. The final loss function was:
\begin{equation}
L=L_{cls}+Dis(\cdot)+Div(\cdot).
\end{equation}
We adopted \textit{Adam}~\cite{adam} as the optimizer with the weight decay of $1e-7$. The $ L_{cls}$ and $( Dis(\cdot)+Div(\cdot))$ were optimized alternately. The learning rate was set to be $1e-3$ for $CGL$, $1e-3$ for $GCN$ and $1e-5$ for backbone during initialization. The models were trained with the same strategy: pretrain the backbone; pretrain the $CGL$; optimize the $CGL$; optimize the $GCN$ for several epochs; optimize the $GCN$ and the backbone; optimize the $CGL$ again if necessary; fine-tune the backbone and the $GCN$ for several epochs until convergence. Please refer to the source code for more details.

\newcommand{\tabincell}[2]{\begin{tabular}{@{}#1@{}}#2\end{tabular}}

\begin{table*}[!thbp]
\centering
\caption{Comparison with the state of the art. The best and second-best results in each column are \textbf{boldfaced} and \underline{underlined}, respectively. ``*'' indicates multi-modal methods that utilize extra user tags, while other methods only utilize the visual information obtained from the images.}


\begin{tabular}{@{} c l l l l l l l l l l  @{}}
\toprule
\multirow{2}[2]{*}{Dataset} & \multirow{2}[2]{*}{Model} & \multirow{2}[2]{*}{Source}& 
\multirow{2}[2]{*}{Accuracy} & \multicolumn{3}{c}{Private} & & \multicolumn{3}{c}{Public}\\
\cmidrule{5-7} \cmidrule{9-11}
& & & & Precision & Recall & F-1 & & Precision & Recall & F-1\\
\midrule
\multirow{5}{*}{PicAlert}  & PCNH & AAAI~\shortcite{tran2016privacy} & 83.15\% & 0.689 & 0.514 & 0.589 & & 0.862 & \underline{0.929} & 0.894\\
& GIP & PR~\shortcite{yang2020graph} & 83.49\% & 0.552 & \underline{0.684} & 0.610 & & \textbf{0.922} & 0.871 & 0.895\\
& *Combination & AAAI~\shortcite{tonge2018uncovering} & 83.09\% & 0.671 & 0.551 & 0.605 & & 0.869 & 0.912 & 0.892\\
& *DMFP & WWW~\shortcite{tonge2019dynamic} & \underline{86.36\%} & \textbf{0.752} & 0.627 & \underline{0.684} & & 0.891 & \textbf{0.936} & \underline{0.913}\\
& DRAG & Ours & \textbf{86.84\%} & \underline{0.719} & \textbf{0.719} & \textbf{0.719} & & \underline{0.914} & 0.914 & \textbf{0.914} \\
\midrule 
Image   & GIP & PR~\shortcite{yang2020graph} & 77.09\% & \textbf{0.812} & 0.751 & 0.780 & & 0.730 & 0.795 & 0.761\\
Privacy& DRAG & Ours & \textbf{87.68\%} & 0.811 & \textbf{0.842} & \textbf{0.826} & & \textbf{0.914} & \textbf{0.895} & \textbf{0.905}\\
\bottomrule
\end{tabular}
\label{exp_result}
\end{table*}

\subsection{Experimental Results}

\subsubsection{Comparison with the State of the Art}
Following previous works, we compared DRAG with state-of-the-art methods on PicAlert and present the precision, recall, and F1 score of each class to validate the effectiveness. We select the models with the best performances on the validation set and report their performances on the test set. The comparisons between the DRAG and the state of the art are presented in Table~\ref{exp_result}.
Note that Combination~\cite{tonge2018uncovering} and DMFP~\cite{tonge2019dynamic} are multi-modal methods that utilize extra user tags, while other methods only utilize the visual information obtained from the images.
As the DRAG only utilizes visual information, we further compare with the state-of-the-art visual-based method~(i.e., GIP) on the more challenging Image Privacy dataset.

We get several observations from the results.
The DRAG outperforms other methods in most metrics on both datasets, which proves the effectiveness of the proposed framework. The accuracy and F-1 score of the DRAG is higher than all other methods, including visual-based and multi-modal ones.
Specifically, the performances in the public class are similar for all methods, and the main difference lies in the private class, which is also the class that we need to pay more attention to.
The DRAG achieved the highest F-1 score and also the highest recall in the private class, which means that the DRAG will significantly reduce the false-negative rate. Considering the practical task of privacy-leaking image detection, this phenomenon means that fewer private images will be classified as public incorrectly, and thus the DRAG will better help reduce the unintentional sharing of private images compared with other methods. 

We observe that the DRAG achieved much better performance compared with GIP, especially on the harder Image Privacy dataset. We analyzed the rationality as follows, and we provide the corresponding Precision-Recall Curve in the supplementary. 
First, as described in the Dataset section,
the public images are the same in the two datasets, while Image Privacy contains more images in the private class. Therefore, Image Privacy is more balanced than PicAlert, and the performances in the public class of both methods dropped on Image Privacy. Second, objects are important clues for privacy-leaking image detection, and thus GIP that relies on the object detector performed well on PicAlert. But when dealing with a more complex dataset, the pretrained object detector limits the ability to focus on other crucial elements like unseen objects, scenes, and textures. 
Compared with GIP, the DRAG dynamically focus on regions of the crucial elements and thus achieved better performances in both classes. 
Our ablation studies in the next subsection also suggest that the model needs to pay more attention to differentiated regions for privacy-leaking image detection on Image Privacy than on PicAlert.

\begin{figure}[t]
\centering
    \subfloat[PicAlert]{%
        \includegraphics[width=0.48\columnwidth]{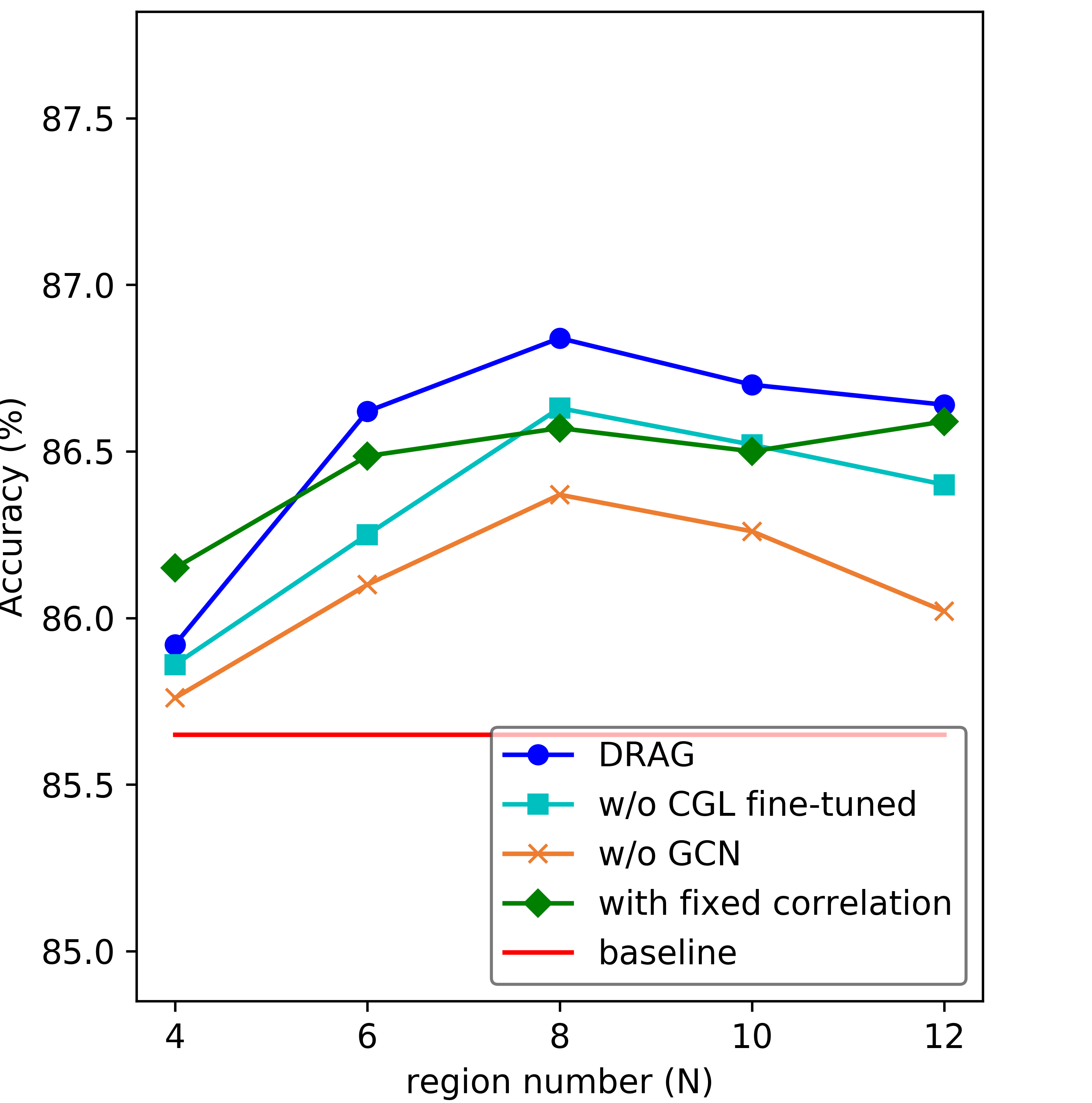}}\hfill
    \subfloat[Image Privacy]{%
        \includegraphics[width=0.48\columnwidth]{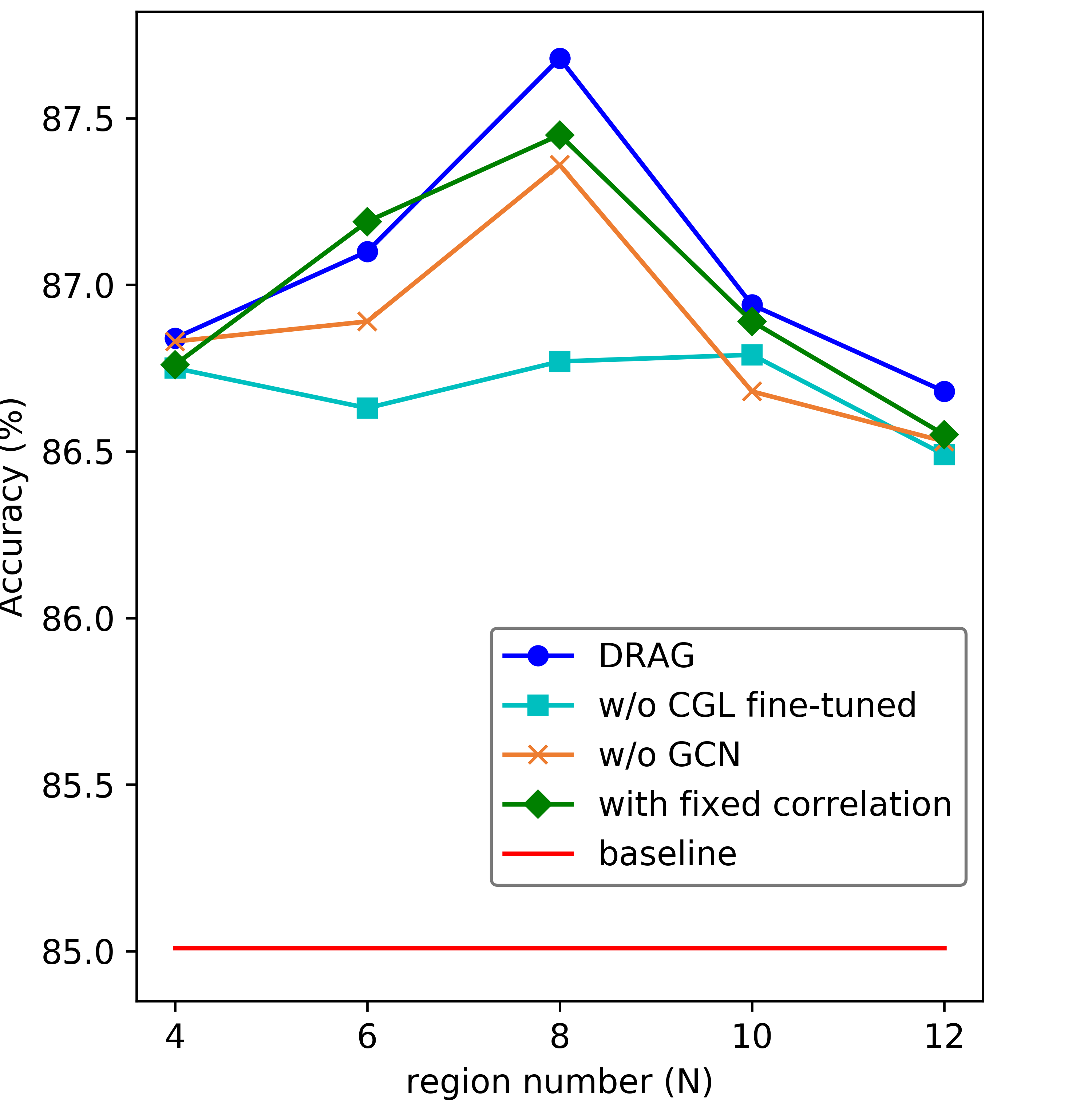}}\hfill

    \caption{Ablation study and hyperparameter sensitivity. The baseline refers to ResNet pretrained on the corresponding dataset (presented in red). The performances drop when removing components from the DRAG, proving the effectiveness of these components. The DRAG is relatively robust for region number $N$, while $N=8$ achieved slightly better performance than other $N$ on both datasets.}
    \label{ablation}
\end{figure}

\subsubsection{Ablation Study}

\ \\
\indent \textbf{Effectiveness of Dynamic Crucial Regions Exploring}
To obtain a variant without the ability to dynamically explore crucial regions, we fixed the $CGL$ with its initial features that mainly focus on objects, because the used backbone was pretrained on the object-focused task.  The results are  presented in cyan in Fig.~\ref{ablation} (``w/o CGL fine-tuned''). The performances drop on both datasets, especially on the more complex Image Privacy. This phenomenon proves that the model needs to explore more elements besides objects, especially for a more complex task.

\textbf{Effectiveness of Dynamic Correlation Modeling}
To obtain a variant without the ability to dynamically model the correlation among crucial regions, we adopted a graph that all nodes in the graph are connected with each other (i.e. a graph with a fixed all-ones correlation matrix). The performances are presented in green in Fig.~\ref{ablation} (``with fixed correlation''), which are worse than those of DRAG in general. 
We further proposed a variant that completely disregards the correlation, which is implemented by removing GCN from DRAG and directly adopts the region features $\mathbf{F_w}$ for classification. Similar to Eq.~\ref{softmax}, we concatenate the features with a global feature, and feed them into a fully connected layer for final prediction: ${\hat{y}} =FC(\rm \mathbf{F_c} \oplus \mathbf{F_w})$. The results are presented in orange in Fig.~\ref{ablation} (``w/o GCN''), and the performances further degrade. 
These results prove that considering the correlation among the crucial elements is essential, while a dynamic correlation is better than a fixed one.

\textbf{Discussion}
For DRAG, the performance is better on Image Privacy than on PicAlert, but for the baseline is the opposite. As described before,
Image Privacy is an extension of PicAlert with challenging samples, which makes the basic model perform worse. The DRAG benefits from the ability to dynamically find out crucial regions and model their correlation, and still achieve remarkable performance.

The dynamic crucial regions exploring and the dynamic correlation modeling complement each other, but the importance of them is different for the two datasets. The dynamic correlation modeling affects the performances more for PicAlert, while the dynamic crucial regions exploring affects the performances more for Image Privacy. 
We speculate that for a simpler dataset, the pretrained $CGL$ is good enough to find out crucial regions, and the model should pay more attention to the interaction among the regions. 
But for a more complex dataset, the model needs to make more effort to focus on the crucial regions that reflect subtle differences. 
This may also explain why GIP that relies on the object detector and GNN performed well on PicAlert, but the performances dropped a lot on Image Privacy.

\begin{figure}[t]
\centering
    \subfloat[Region features of a private image]{%
        \includegraphics[width=1\columnwidth]{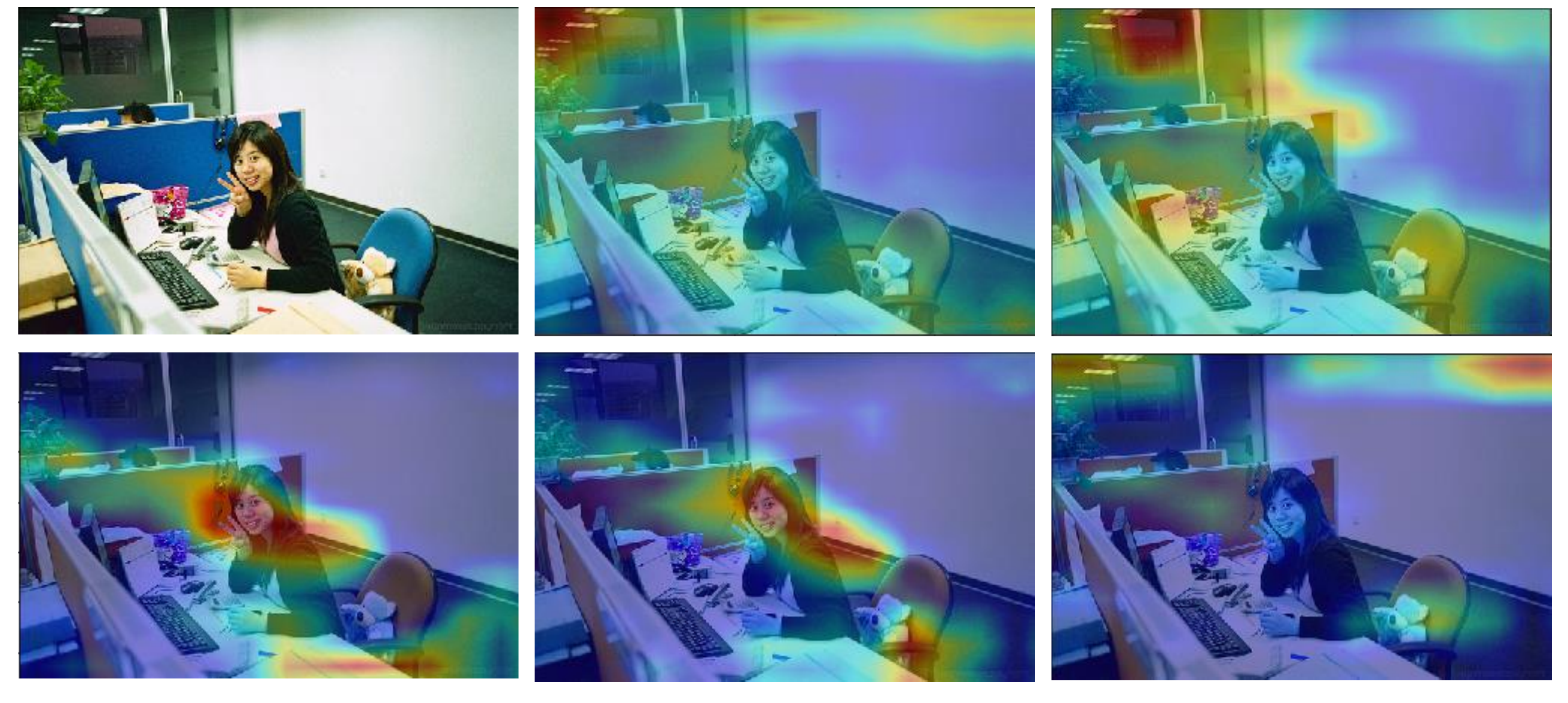}}\hfill
    \subfloat[Region features of a public image]{%
        \includegraphics[width=1\columnwidth]{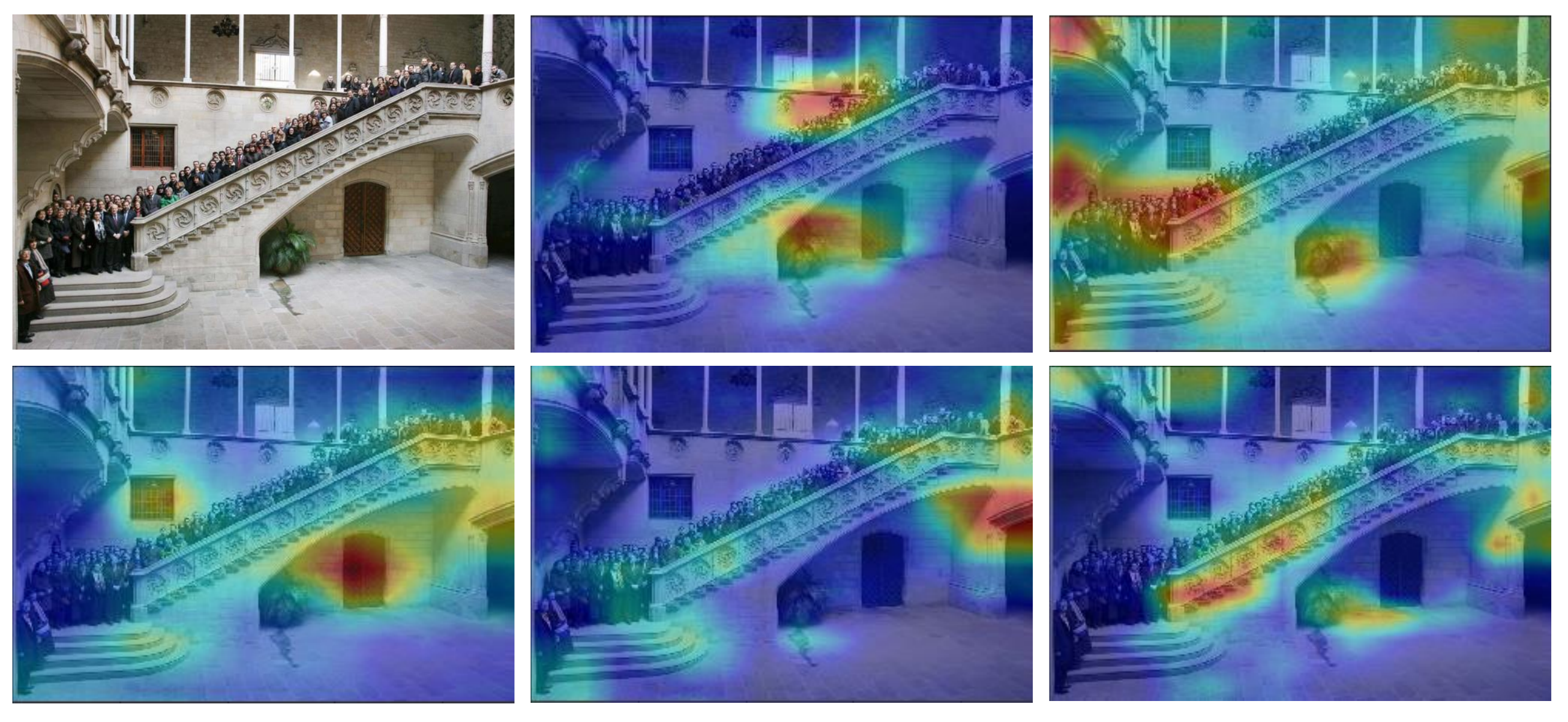}}\hfill

    \caption{Two examples of the learned region feature obtained from $CGL$. In the private image (a), the $CGL$ focuses on the lamp, the window, the person, the doll on the chair, and the wall. In the public image (b), the $CGL$ focuses on the plant, the people, the doors, and the railing. The results show that the $CGL$ can capture regions of crucial elements to differentiate private and public images.}
    \label{visualization}
\end{figure}

\subsubsection{Hyperparameter Sensitivity}
To validate the robustness of the model under parameter variations, we investigated the sensitivity of the hyperparameter $N$, which determines the number of regions during channel grouping, and also the number of nodes in the GCN. 
The results are presented in Fig.~\ref{ablation}. We explore the influence of ${N}$ with the complete model as well as the models in the ablation studies. The tendencies are consistent for all models, and we can get several conclusions. 
First, for most models, the variances of the performances between different ${N}$ are not very significant, suggesting that the DRAG is relatively robust. 
Second, comparing the subtle differences
, we found that the region number of $8$ is most suitable for our experimental setups on both datasets. During our experiments, we infer that the best selection of $N$ may depend on the size of the feature map $\mathbf{F_b}$ used for clustering. A larger $N$ may be more appropriate for a larger feature map. We will explain this inference based on the following case study.

\subsection{Case study}
\textbf{Qualitative Analyses of the Region Features  }
\label{cases}
We visualized the original image and corresponding 
features of the crucial regions obtained from $CGL$ to illustrate the capability of $CGL$. 
From Fig.~\ref{visualization}, we observe that the $CGL$ learned differentiated regions of crucial elements as we expected. We also notice that there exist overlaps between the peak responses in the feature maps, and several feature maps are not very compact. During the training stage, we combined the losses to ensure the classification performance and did not make a quite strict constraint on the $Dis(\cdot)$ and $Div(\cdot)$, and thus the results are reasonable. 
This may also explain why the region number $N$ of $8$ is the most suitable one in our experiments --- too small $N$ may neglect important region features, while too large $N$ may result in overlaps.

\begin{figure}[!htbp]
\centering
    \subfloat[Public images that are misclassified into private.]{%
        \includegraphics[width=0.49\columnwidth, height=1.5in]{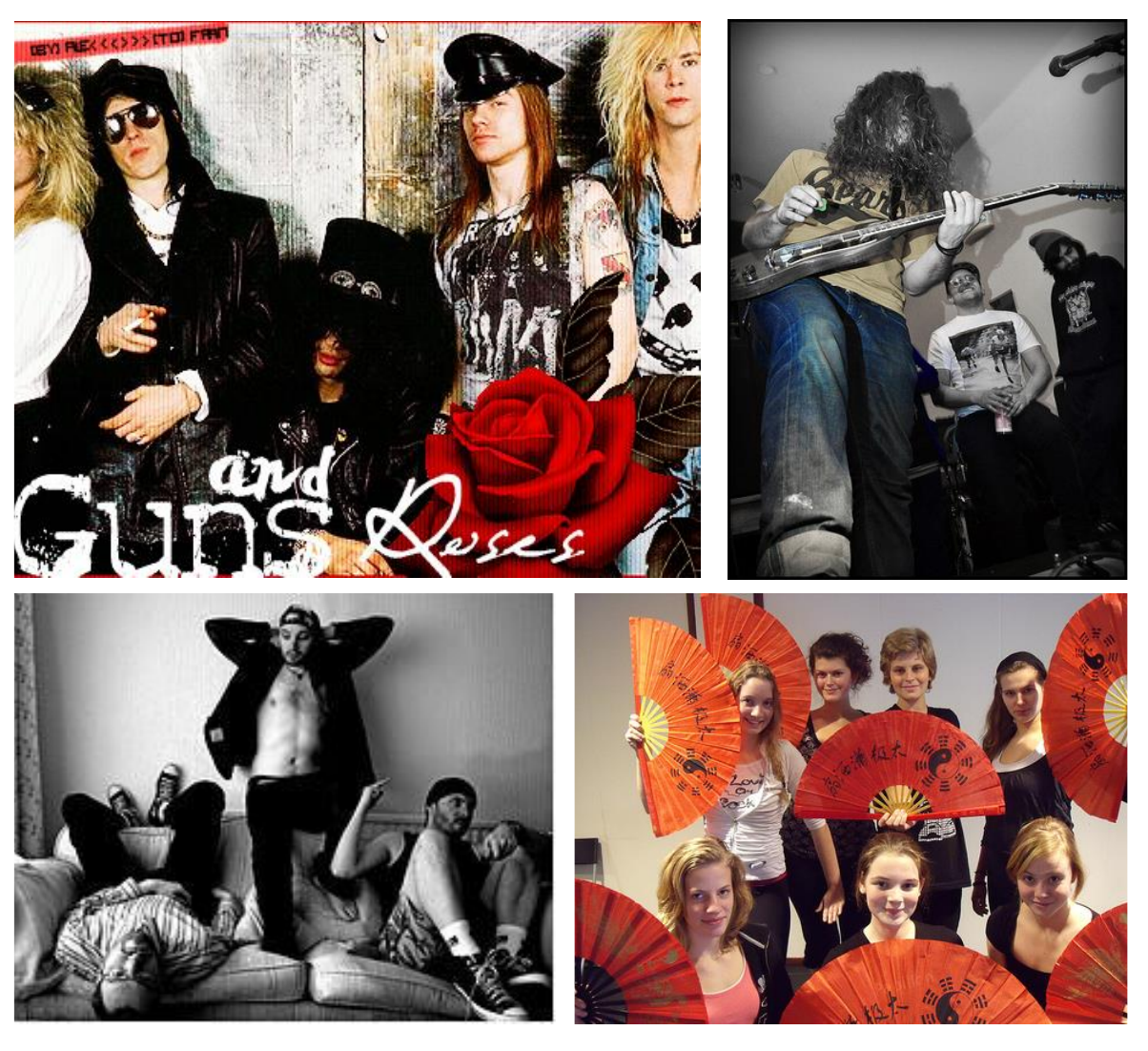}}\hfill
    \subfloat[Private images that are misclassified into public.]{%
        \includegraphics[width=0.49\columnwidth,height=1.5in]{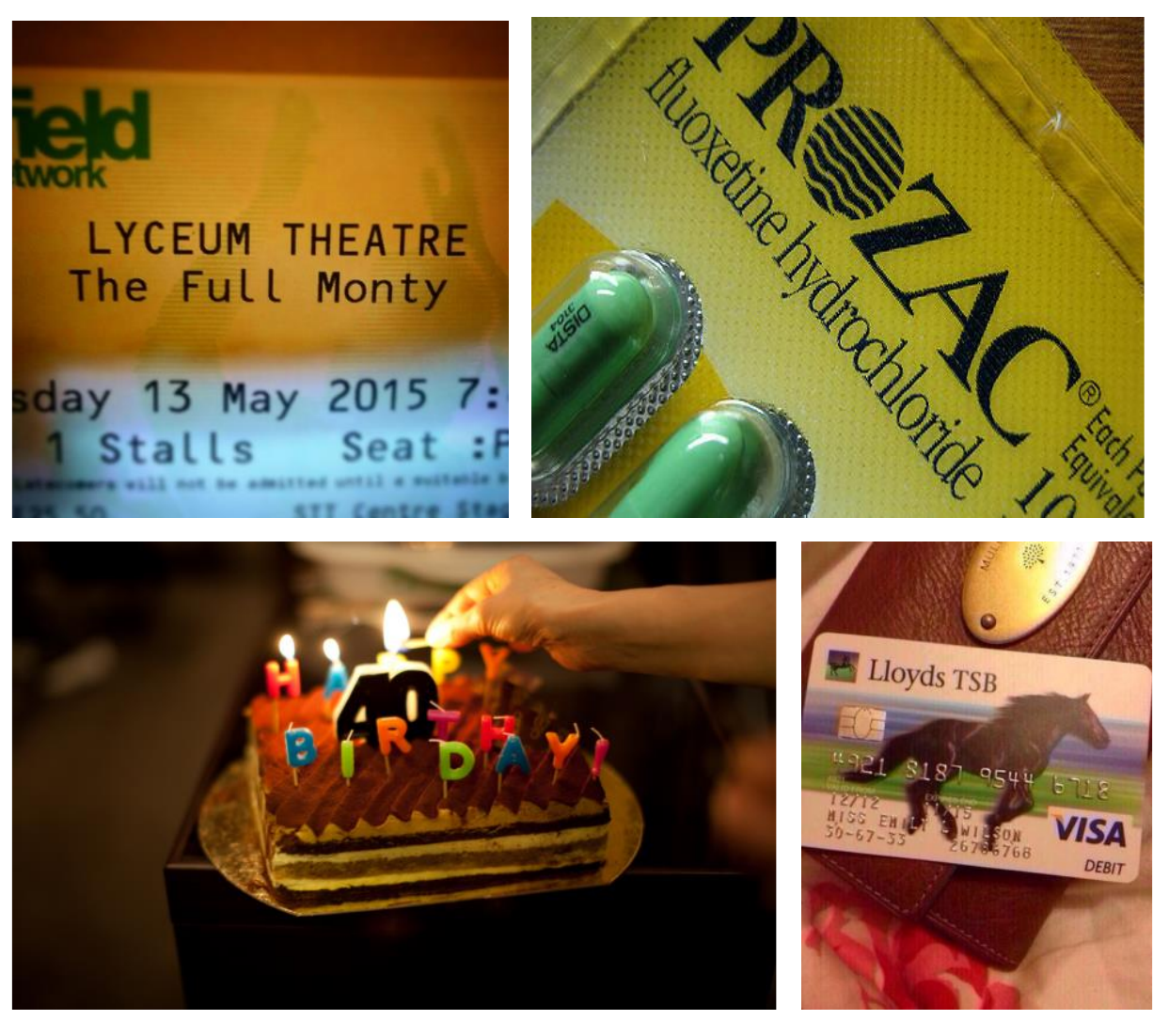}}\hfill

    \caption{Cases of misclassified images.
    }
    \label{bad_cases}
\end{figure}

\noindent\textbf{Limitation of DRAG  }
We conduct this study to learn what kind of images are more likely to be misclassified. Fig.~\ref{bad_cases} (a) shows public images that were misclassified into \textit{private}. Although group photos are often related to private occasions like family gatherings, the images here are actually art photography and photos of public events. In Fig.~\ref{bad_cases} (b), the misclassified private images contain elements such as ticket, medicine, age, and credit card number. The examples indicate that the model may fail to understand the given images when the external social context of sharing motivation and textual privacy is necessary. Therefore, we argue that future works may obtain a deeper understanding of the images by introducing social context. For example, techniques of text recognition and natural language understanding can be used to understand the specific types of card-like elements.

\section{Conclusion}
In this paper, we proposed the DRAG for privacy-leaking image detection. The DRAG dynamically finds out crucial regions and models their correlation adaptively for each input image without the limitation of pretrained object detectors. The experimental results show that the DRAG that only utilizes visual features outperformed existing methods, including visual-based and multi-modal ones. Further works may consider introducing external social context to obtain a deeper understanding of the images. The code will be released to facilitate further research.

\section{Acknowledgements}
The corresponding author is Juan Cao. This work was supported by the Zhejiang Provincial Key Research and Development Program of China (NO. 2021C01164), the Project of Chinese Academy of Sciences (E141020), and the National Natural Science Foundation of China~(No. 62172420). 

The authors thank Wu Liu, Xinchen Liu, Tianyun Yang, Lei Li, and anonymous reviewers for their helpful advice on the paper. We also thank Yanyan Wang and Lei Zhong for their help on the implementation of the model. 

\bibliography{image_privacy}

\end{document}